\pdfoutput=1

\documentclass[11pt]{article}

\usepackage[]{acl}

\usepackage{times}
\usepackage{latexsym}
\usepackage{cuted}

\usepackage[T1]{fontenc}

\usepackage[utf8]{inputenc}

\usepackage{microtype}
\usepackage{xspace}
\usepackage{xcolor}
\usepackage{booktabs}
\usepackage{array,multirow,graphicx}
\usepackage{makecell}
\usepackage{color, colortbl}
\usepackage{amssymb}
\usepackage{amsmath,amsfonts}
\usepackage{amsopn}
\usepackage{bm} 
\usepackage{multirow}
\usepackage{tabularx}

\newcommand{\ProbOpr}[1]{\mathbb{#1}}

\newcommand{\expect}[2]{%
\ifthenelse{\equal{#2}{}}{\ProbOpr{E}_{#1}}
{\ifthenelse{\equal{#1}{}}{\ProbOpr{E}\left[#2\right]}{\ProbOpr{E}_{#1}\left[#2\right]}}} 
\newcommand{\var}[2]{%
\ifthenelse{\equal{#2}{}}{\ProbOpr{VAR}_{#1}}
{\ifthenelse{\equal{#1}{}}{\ProbOpr{VAR}\left[#2\right]}{\ProbOpr{VAR}_{#1}\left[#2\right]}}} 

\newcommand{\eat}[1]{}

\newcommand{\eg}{{\em e.g.}}
\newcommand{\ie}{{\em i.e.}}

\usepackage{amssymb}
\usepackage{pifont}
\newcommand{\cmark}{\ding{51}}%
\newcommand{\xmark}{\ding{55}}%
\definecolor{Gray}{gray}{0.6}
\definecolor{LightCyan}{rgb}{0.88,1,1}
\newlength\savewidth
\newcommand\mypara[1]{\vspace{1.0mm}\noindent\textbf{#1}}
\newcommand{\ourmethod}{{\sc {SimpleAug}}\xspace}
\newcommand{\ourmethodbf}{{\textbf{\textsc {SimpleAug}}}\xspace}

\usepackage{enumitem}

\newcommand*{\rowstyle}[1]{
	\gdef\@rowstyle{#1}
	\@rowstyle\ignorespaces%
}
\newcolumntype{=}{
	>{\gdef\@rowstyle{}}
}
\newcolumntype{+}{
	>{\@rowstyle}
}

\title{Discovering the Unknown Knowns: \\ Turning Implicit Knowledge in the Dataset into Explicit Training Examples for Visual Question Answering}

\author{Jihyung Kil, \hfill Cheng Zhang, \hfill Dong Xuan, \hfill Wei-Lun Chao \\
The Ohio State University, Columbus, OH, USA \\
\small \texttt{\{kil.5, zhang.7804, xuan.3, chao.209\}@osu.edu}
}

\begin{document}
\maketitle


\begin{abstract}
Visual question answering (VQA) is challenging not only because the model has to handle multi-modal information, but also because it is just so hard to collect sufficient training examples --- there are too many questions one can ask about an image. As a result, a VQA model trained solely on human-annotated examples could easily over-fit specific question styles or image contents that are being asked, leaving the model largely ignorant about the sheer diversity of questions. Existing methods address this issue primarily by introducing an auxiliary task such as visual grounding, cycle consistency, or debiasing. In this paper, we take a drastically different approach. We found that many of the ``unknowns'' to the learned VQA model are indeed ``known'' in the dataset implicitly.
For instance, questions asking about the same object in different images are likely paraphrases; the number of detected or annotated objects in an image already provides the answer to the ``how many'' question, even if the question has not been annotated for that image. Building upon these insights, we present a simple data augmentation pipeline \ourmethod to turn this ``known'' knowledge into training examples for VQA. We show that these augmented examples can notably improve the learned VQA models' performance, not only on the VQA-CP dataset with language prior shifts but also on the VQA v2 dataset without such shifts. Our method further opens up the door to leverage weakly-labeled or unlabeled images in a principled way to enhance VQA models. Our code and data are publicly available at \url{https://github.com/heendung/simpleAUG}.
\end{abstract}


\section{Introduction}
\label{s_intro} 

``A picture is worth a thousand words,'' which tells how expressive an image can be, but also how challenging it is to teach a machine to understand an image like we humans do. 
Visual question answering (VQA)~\cite{antol2015vqa,goyal2017making,zhu2016visual7w} is a principled way to measure such an ability of a machine, in which given an image, a machine has to answer the image-related questions in natural language by natural language.
While after years of effort, the state-of-the-art machine's performance is still behind what we expect~\cite{hu2018learning,yu2018beyond,anderson2018bottom,lu2019vilbert,hudson2019gqa}.

\begin{figure}[t]
    \centerline{\includegraphics[width=1\linewidth]{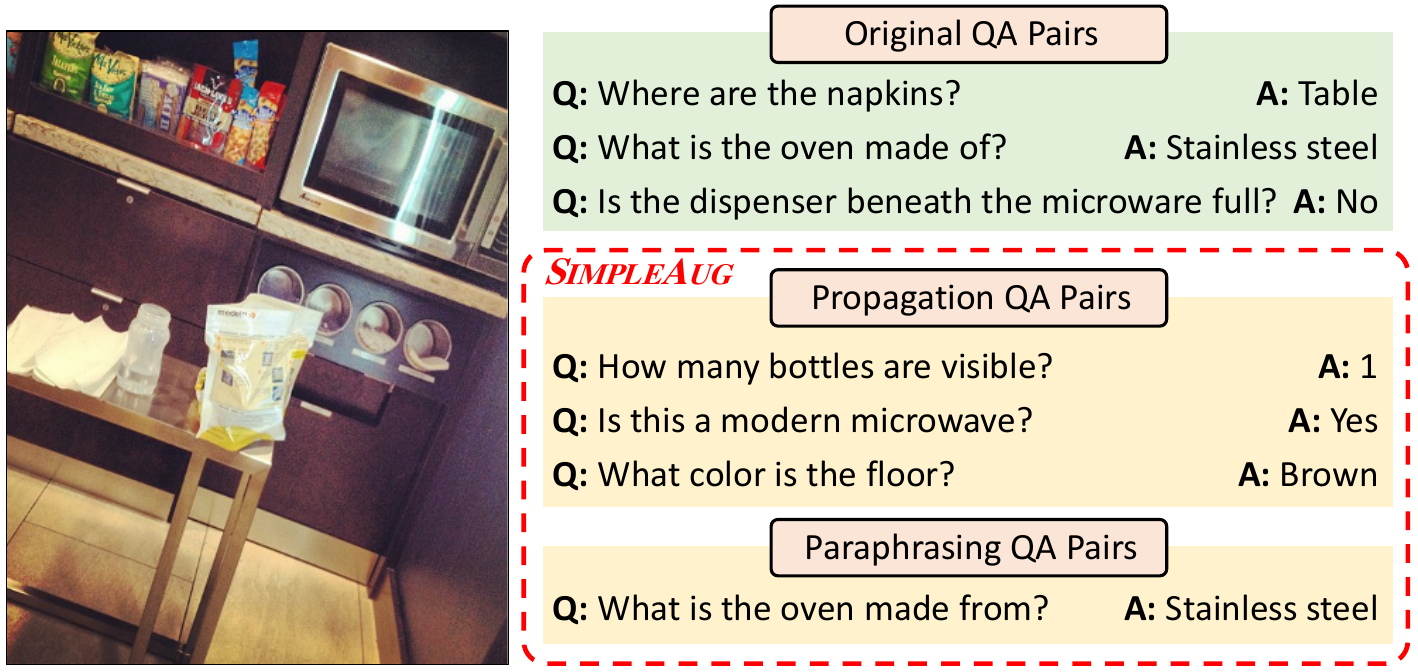}}
    \caption{\small \textbf{Illustration of our approach \ourmethodbf}. We show a \emph{training} image and its corresponding question-answer pairs in VQA v2~\cite{goyal2017making}, and our generated pairs. A VQA model~\cite{anderson2018bottom} trained on the original dataset \emph{just cannot answer these new questions on the training image correctly}, and we use them to improve model training.}
    \label{fig:1}
    \vskip -5pt
\end{figure}

Several key bottlenecks have been identified. In particular, a machine (\ie, VQA model) learned in the conventional supervised manner using human-annotated image-question-answer (IQA) triplets is shown to overlook the image or language contents~\cite{agrawal2016analyzing,goyal2017making,chao2017being}, over-fit the language bias~\cite{agrawal2018don}, or struggle in capturing the diversity of human language~\cite{shah2019cycle,chao2018cross}. Many recent works thus propose to augment the original VQA task with auxiliary tasks or losses such as visual grounding~\cite{selvaraju2019taking,wu2019self}, de-biasing~\cite{cadene2019rubi,clark2019don,ramakrishnan2018overcoming}, or (cycle-)consistency~\cite{shah2019cycle,gokhale2020mutant} to address these issues.

Intrigued by these findings and solutions, we investigate the bottlenecks further and argue that they may result from a more fundamental issue --- there are simply not enough training examples (\ie, IQA triplets). Concretely, most of the existing VQA datasets annotate each image with around ten questions, which are much fewer than what we humans can ask about an image. Take the popular VQA v2 dataset~\cite{goyal2017making}, for instance, the trained VQA model can answer most of the training examples (on average, six questions per image) correctly. However, if  we ask some more questions about the training images --- \eg, by borrowing relevant questions from other training images --- the same VQA model fails drastically, \emph{even if the model has indeed seen these images and questions during training} (see \autoref{fig:1}). Namely, the VQA model just has not learned enough through the human-annotated examples, leaving the model unaware of the huge amount of visual information in an image and how it can be asked via natural language.

At first glance, this seems to paint a grim picture for VQA.
However, in this paper we propose to take advantage of this weakness to strengthen the VQA model: we turn implicit information already in the dataset, such as unique questions and the rich contents in the training images, into explicit IQA triplets which can be directly used by VQA models via conventional supervised learning.

We propose a simple data augmentation method \ourmethod, which relies on (i) the original image-question-answer triplets in the dataset, (ii) mid-level semantic annotations available on the training images (\eg, object bounding boxes),
and (iii) pre-trained object detectors~\cite{ren2016faster}\footnote{We note that (ii) is commonly provided in existing VQA datasets like VQA v2~\cite{goyal2017making}, and (iii) has been widely used in the feature extraction stage of a VQA model~\cite{anderson2018bottom}.}. Concretely, we build upon the aforementioned observations --- \emph{questions annotated for one image can be valuable add-ons to other relevant images} --- and design a series of mechanisms to ``propagate'' questions from one image to the others. More specifically, we search images that contain objects mentioned in the question and identify the answers using information provided by (ii) and (iii), such as numbers of objects, their attributes, and existences.

\ourmethod requires no question generation step via templates or language models~\cite{kafle2017data}, bypassing the problems of limited diversity or artifacts.
Besides, \ourmethod is completely detached from the training phase of a VQA model and is therefore model-agnostic, making it fairly simple to use to improve VQA models.

We validate \ourmethod on two datasets, VQA v2~\cite{goyal2017making} and VQA-CP~\cite{agrawal2018don}. The latter is designed to evaluate VQA models' generalizability under language bias shifts. With \ourmethod, we can not only achieve comparable gains to other existing methods on VQA-CP, but also boost the accuracy on VQA v2, demonstrating the applicability of our method. We note that many of the prior works designed for VQA-CP indeed degrade the accuracy on VQA v2, which does not have language bias shifts between training and test data. \ourmethod further justifies that mid-level vision tasks like object detection can effectively benefit high-level vision tasks like VQA.

In summary, our contributions are three-folded:

\begin{itemize} [itemsep=0pt,topsep=0pt,leftmargin=10pt]
    \item We propose \ourmethod, a simple and model-agnostic data augmentation method that turns information already in the datasets into explicit IQA triplets for training VQA models.
    \item We show that \ourmethod can notably improve VQA models' accuracy on both VQA v2~\cite{goyal2017making} and VQA-CP~\cite{clark2019don}.
    \item We provide comprehensive analyses on \ourmethod, including its applicability to
    weakly-labeled and unlabeled images.
\end{itemize}


\section{Related Work}
\label{s_related}

\mypara{VQA datasets.} 
More than a dozen datasets have been released~\cite{lin2014microsoft,antol2015vqa,zhu2016visual7w,goyal2017making,krishna2017visual,hudson2019gqa,gurari2019vizwiz}.
Most of them use natural images from large-scale image databases, \eg, MSCOCO~\cite{lin2014microsoft}. 
For each image, human annotators are asked to generate questions
(Q) and provide the corresponding answers (A). Doing so, however, is hard to cover all the knowledge in the visual contents.

\mypara{Leveraging side information for VQA.}
A variety of side information beyond the IQA triplets has been used to improve VQA models. For example, human attentions are used to enhance the explainability and visual grounding of VQA models~\cite{patro2018differential,selvaraju2019taking,das2017human,wu2019self}.
Image captions contain substantial visual information and can be used as an auxiliary task (\ie, visual captioning) to strengthen VQA models' visual and language understanding
\cite{wu2019generating,kim2019improving,wang2021latent,banerjee2020self,karpathy2015deep}.
Several papers leveraged scene graphs and visual relationships as auxiliary knowledge for VQA~\cite{johnson2017inferring,zhang2019empirical,hudson2019gqa,shi2018explainable}.
A few works utilized mid-level vision tasks (\eg, object detection and segmentation) to benefit VQA~\cite{gan2017vqs,kafle2017data}.
Most of these works use side information by defining auxiliary learning tasks or losses to the original VQA task.
In contrast, we directly turn the information into IQA triplets for training. 

\mypara{Data augmentation for VQA.}
Several existing works investigate data augmentation.
One stream of works creates new triplets by manipulating images or questions~\cite{chen2020counterfactual,agarwal2020towards,tang2020semantic,gokhale2020mutant}. See~\S~\ref{ss_baseline} for some more details.
The other creates more questions by using a learned language model to paraphrase sentences~\cite{ray2019sunny,shah2019cycle,whitehead2020learning,kant2020contrast,banerjee2020self} or by learning a visual question generation model~\cite{kafle2017data,li2018visual,krishna2019information}.

The closest to ours is the pioneer work of data augmentation by~\citet{kafle2017data}, which also creates new questions by using mid-level semantic information annotated by humans (\eg, object bounding boxes).
Their question generation relies on either pre-defined templates or a learned language model, which may suffer limited diversity or labeling noise. In contrast, we directly reuse questions already in the dataset and show that they are sufficient to augment high-quality questions for other images.
Besides, we further explore machine generated annotations (\eg, via an object detector~\cite{ren2016faster}), opening the door to augment triplets using extra unlabeled images.
Furthermore, we benchmark our method on the popular VQA v2~\cite{goyal2017making} and challenging VQA-CP~\cite{agrawal2018don} datasets, which are released after the publication of \citet{kafle2017data}.
\emph{Overall, we view our paper as an attempt to revisit simple data augmentation like \citet{kafle2017data} for VQA, and show that it is indeed quite effective.} 

\mypara{Robustness of VQA models.}
\citet{goyal2017making,chao2017being,agrawal2018don} pointed out the existence of superficial correlations (\eg, language bias) in the datasets and showed that 
a VQA model can simply exploit them to answer questions. 
Existing works to address this can be categorized into three groups. The first group attempts to reduce the language bias by designing new VQA models or learning strategies~\cite{agrawal2018don,ramakrishnan2018overcoming,cadene2019rubi,clark2019don,grand2019adversarial,clark2019don,jing2020overcoming,niu2020counterfactual,shrestha2020negative,gat2020removing}.
For example,  
RUBi and Ensemble~\cite{cadene2019rubi, clark2019don} explicitly modeled the question-answer correlations to encourage VQA models to explore other patterns in the data that are more likely to generalize. 
The second group leverages side information to facilitate visual grounding~\cite{wu2019self,selvaraju2019taking,teney2019actively}. For example, 
\citet{wu2019self} used extra visual or textual annotations to determine important regions where a VQA model should focus on. 
The third group implicitly or explicitly augments the VQA datasets, \eg, via self-supervised learning, counterfactual sampling, {adversarial training, or image and question manipulation~\cite{abbasnejad2020counterfactual,zhu2020overcoming,teney2020value,chen2020counterfactual,gokhale2020mutant,liang2020learning,gokhale2020vqa,li2020closer,ribeiro2019red,selvaraju2020squinting}.}
\ourmethod belongs to the third group but is simpler in terms of methodology. Besides, \ourmethod is completely detached from VQA model training and thus model-agnostic. Moreover, \ourmethod can improve on both VQA-CP~\cite{agrawal2018don} and VQA v2~\cite{goyal2017making}.
%

\section{\ourmethodbf for Data Augmentation}
\label{s_approach}

\subsection{Implicit information} 

\begin{figure*}[h]
    \centerline{\includegraphics[width=\linewidth]{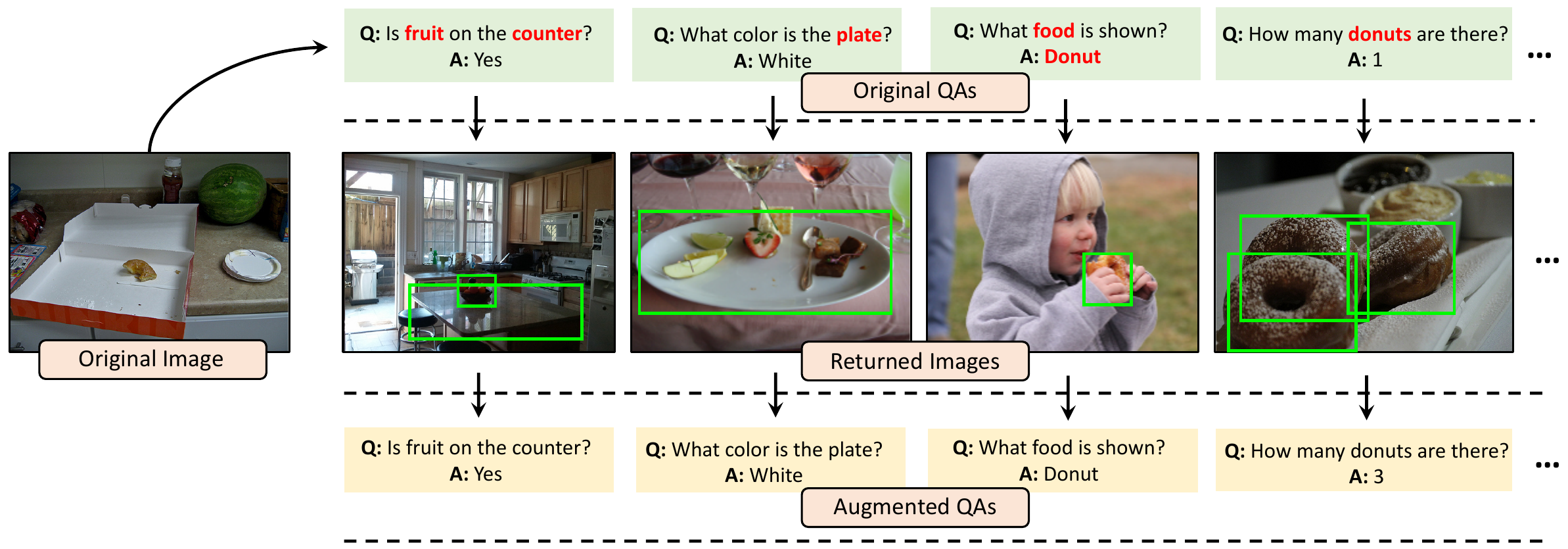}}
    \caption{\small \textbf{The \ourmethodbf pipeline.} We show four original question-answer pairs of the image on the left in VQA v2, and how they are propagated to other images. The {\color{green}green} boxes are annotated in MSCOCO or detected by Faster R-CNN; each of them is associated with an object name and/or attribute. We only show boxes matched by {\color{red} nouns} or used to derive answers.
    }
    \label{fig:pipeline}
    \vskip -5pt
\end{figure*}

\ourmethod leverages three sources of information that implicitly suggest extra IQA triplets beyond those provided in a VQA dataset. The \textbf{first} one is the original IQA triplets in the dataset. We find that for two similar images that locally share common objects or globally share common layouts, their corresponding annotated questions can either be treated as paraphrases or extra questions for each other. The \textbf{second} one is the object instance labels like object bounding boxes that are annotated on images, \eg, MSCOCO images~\cite{lin2014microsoft}. These labels provide accurate answers to ``how many'' or some of the ``what'' questions, and many VQA datasets are built upon MSCOCO images.
The \textbf{third} one is an object detector pre-trained on an external densely-annotated dataset like Visual Genome (VG)~\cite{krishna2017visual}. This detector can provide information not commonly annotated on images, such as attributes or fine-grained class names. We note that since the seminal work by \citet{anderson2018bottom}, many following-up VQA models use the Faster R-CNN detector~\cite{ren2016faster} pre-trained on VG for feature extraction. 

\subsection{The \ourmethodbf pipeline}
\label{ss_simple_aug_2}

\ourmethod processes each annotated IQA triplet $(i,q,a)$ in term, and propagates $q$ to other relevant images. 
To begin with, \ourmethod extract meaningful words from the question, similar to~\cite{chen2020counterfactual,wu2019self}. We leverage a spaCy part-of-speech (POS) tagger~\cite{honnibal2017spacy} to extract ``nouns'' and tokenize their singular and plural forms. 
We remove words such as ``picture'' or ``photo'', which appear in many questions but are not informative for VQA\footnote{For example, in a question, ``What is the person doing in the picture?'', the word ``picture'' refers to the image itself, not an object within it. We found 8\% of the questions like this in VQA v2~\cite{goyal2017making}. Some questions really refer to ``pictures'' or ``photos'' within an image (\eg, ``How many pictures on the wall?''), but there are $<$1\% such questions.}.

Given the meaningful ``nouns'' of a question, we then retrieve relevant images and derive the answers, using MSCOCO annotations or Faster R-CNN detection.
Concretely, we split questions into four categories and develop specific \emph{question propagation rules}.
\autoref{fig:pipeline} illustrates the pipeline.

\mypara{Yes/No questions.}
We apply the Faster R-CNN detector trained on VG to each image $i'$ beside image $i$. The detector returns a set of bounding boxes and their labels. We ignore images whose object labels have no overlap with the nouns of question $q$, and assign answer ``yes'' or ``no'' to the remaining images as follows.

\begin{itemize} [itemsep=0pt,topsep=0pt,leftmargin=10pt]
    \item ``Yes'': if the labels of image $i'$ cover all nouns of question $q$, we create $(i', q, yes)$.
    \item ``No'': if the labels of image $i'$ only cover some of the nouns of $q$, we create $(i', q, no)$. For instance, if the question is ``Is there a cat on the pillow?'' but the image only contains ``pillow'' but no ``cat'', then the answer is ``no''.
\end{itemize}

We develop two verification strategies at the end of this subsection to filter out outlier cases.

\mypara{Color questions.} To prevent ambiguous cases, we only consider questions with a single noun (besides the word ``color'').
We again apply the Faster R-CNN detector, which returns for each image a set of object labels that may also contain attributes like colors. We keep images whose labels cover the noun of question $q$. For each such image $i'$, we create a triplet $(i', q, \hat{a})$, where $\hat{a}$ is the color attribute provided by the detector. As there are likely some other object-color pairs in $i'$, we investigate replacing the noun in $q$ by each detected object name and create some more IQA triplets about colors.

\mypara{Number questions.} We again focus on questions with a single noun (besides the word ``number'').
We use MSCOCO annotations, which give each image a set of object bounding boxes and labels. We find all the images whose labels cover the noun of question $q$. For each such image $i'$, we derive the answer by counting annotated instances of that noun and create a triplet $(i', q, \hat{a})$, where $\hat{a}$ is the count. Some of the nouns (\eg, ``animal'') are super-categories of sub-category objects (\eg, ``dog'' or ``cat''). Thus, if the noun of $q$ is a super-category (\eg, $q$ is ``How many animals are there?''), we follow the category hierarchy provided by MSCOCO and 
count all its sub-category instances.

\mypara{Other questions.} We focus on ``what'' questions with a single noun and use MSCOCO annotations. We find all the images whose labels cover the noun of question $q$. (We also take the super-category cases into account.)
For each such image $i'$, we check whether its MSCOCO labels contain the answer $a$ of question $q$, \ie, according to the original $(i,q,a)$ triplet. For instance, if $q$ is ``What animal is this?'' and $a$ is ``sheep'', then we check if image $i'$'s labels contain ``sheep''. If yes, we create a triplet $(i', q, \hat{a} = a)$. This process essentially discovers ``what'' questions that can indeed be asked about $i'$.

\mypara{Verification.}
The above rules simplify a question by only looking at its nouns, so they may lead to triplets whose answers are incorrect. To mitigate this issue, we develop two verification strategies.

The first strategy performs self-verification on the original $(i, q, a)$ triplet, checking if our rules can reproduce it. That is, it applies the aforementioned rules to image $i$ to derive the new triplet $(i, q, \hat{a})$. If $\hat{a}$ does not match $a$, \ie, using the rules creates a different answer, we skip this question $q$.

The second strategy verifies our rules using IQA triplets annotated on retrieved image $i'$. For example, if image $i'$ has an annotate triplet $(i', q', a')$ whose $q'$ has the same category and nouns as $q$, then we compare it to the created triplet $(i', q, \hat{a})$. If $\hat{a}$ does not match $a'$, then we disregard $(i', q, \hat{a})$.

\subsection{Paraphrasing by similar questions}
\label{ss_para}
Besides the four question propagation rules that look at the image contents, we also investigate a simple paraphrasing rule by searching similar questions in the dataset. Concretely, we apply the averaged word feature from BERT~\cite{devlin2018bert} to encode each question as it better captures the object-level semantics for searching questions mentioning the same objects. Two questions are similar if their cosine similarity is above a certain threshold ($0.98$ in the experiments). If two IQA triplets $(i, q, a)$ and $(i', q', a')$ have similar questions, we create two extra triplets $(i, q', a)$ and $(i', q, a')$ by switching their questions as paraphrasing. 

We choose a high threshold 0.98 to avoid false positives.
On average, each question finds 11.4 similar questions, and we only pick the top-3 questions. We found that with this design, an extra verification step, like checking if the image $i'$ contains nouns of the paraphrasing question $q$, does not further improve the overall VQA accuracy. Thus, we do not include an extra verification step for paraphrasing.

\section{Experiments}
\label{s_exp}

\subsection{Experimental setup}
\label{ss_setup}

\subsubsection{VQA datasets and evaluation metrics}
We validate \ourmethod on two popular datasets. {See Appendix for the results on other datasets.}

\mypara{VQA v2}~\cite{goyal2017making} collects images from MSCOCO~\cite{lin2014microsoft} and uses the same training/validation/testing splits. 
On average, six questions are annotated for each image. In total, VQA v2 has 
444K/214K/448K
training/validation/test IQA triplets. 

\mypara{VQA-CP v2}~\cite{agrawal2018don} is a challenging \emph{adversarial} split of VQA v2 designed to evaluate the model's capability of handling language bias/prior shifts between training and testing. For instance, ``white'' is the most frequent answer for questions that start with ``what color...'' in the training set whereas ``black'' is the most common one in the test set. Such \emph{prior changes} also reflect in individual questions, \eg, the most common answer for ``What color is the banana?'' changes from ``yellow'' during training to ``green'' during testing. VQA-CP v2 has 438K/220K
training/test IQA triplets.

\mypara{Evaluation metrics.} We follow the standard evaluation protocol~\cite{antol2015vqa,goyal2017making}. For each test triplet, the predicted answer 
is compared with answers 
provided by ten human annotators in a leave-one-annotator-out fashion for robust evaluation. We report the averaged scores over all test triplets as well as over test triplets of Yes/No, number, or other answer types.
\subsubsection{Implicit knowledge sources}
\mypara{MSCOCO annotations}~\cite{lin2014microsoft}.
MSCOCO is the most popular benchmark nowadays for object detection and instance segmentation, which contains 80 categories (\eg, ``cat'') as well as the corresponding super categories (\eg, ``animal''). Object instances of all 80 categories are exhaustively annotated in all images, leading to approximately 1.2 million instance annotations.
\mypara{Faster R-CNN detection}~\cite{anderson2018bottom}.
We use the object detection results from a Faster R-CNN~\cite{ren2016faster} pre-trained with Visual Genome (VG)~\cite{krishna2017visual}. This pre-trained detector can provide object attributes (\eg, color and material) whereas MSCOCO annotations only contain object names (\eg, ``person'' and ``bicycle''). We use the detector provided by~\citet{anderson2018bottom}, which detects 36 objects per image.

\subsubsection{Base VQA models}
\ourmethod is model-agnostic, and we evaluate it by using its generated data to augment the training set for training
three base VQA models.

\mypara{Bottom-Up Top-Down (UpDn)}~\cite{anderson2018bottom}. 
UpDn is a widely used VQA model. It first detects objects from an image and encodes them into visual feature vectors. 
Given a question, UpDn uses a question encoder to produce a set of word features. Both visual and language features are then fed into a multi-modal attention network to predict the answer.

\mypara{Learned-Mixin+H (LMH)}~\cite{clark2019don}.
LMH is a learning strategy to de-bias a VQA model, \eg, UpDn. During training, LMH uses an auxiliary question-only model to encourage the VQA model to explore visual-question related information. During testing, only the VQA model is used. LMH is shown to largely improve the performance on VQA-CP v2 but can hurt that on VQA v2. 
\mypara{LXMERT}~\cite{tan2019lxmert}.
{We also study \ourmethod with a stronger, transformer-based VQA model named LXMERT. 
LXMERT leverages multi-modal transformers to extract multi-modal features, and  exploits a masking mechanism to better (pre-)train the model.
While such a masking mechanism can be viewed as a way of data augmentation, \ourmethod is fundamentally different from it in two aspects.
First, \ourmethod generates new triplets while masking manipulates existing triplets. Second, \ourmethod is detached from model training and is therefore compatible with masking.} As will be shown in the experimental results, \ourmethod can provide solid gains to LXMERT on both VQA v2 and VQA-CP.

\subsubsection{Compared data augmentation methods}
\label{ss_baseline}
We compare \ourmethod with three 
existing data augmentation methods for VQA.

\mypara{Template-based} augmentation proposed by~\citet{kafle2017data} generates new question-answer pairs using MSCOCO annotations (cf. \S~\ref{s_related}). We re-implement the method following the paper.  

\mypara{Counterfactual Samples Synthesizing (CSS)} 
\cite{chen2020counterfactual} generates counterfactual triplets by masking critical objects in images or words in questions and assigning different answers.
These new training examples force the VQA model to focus on those critical objects and words, improving both visual explainability and question sensitivity.

\mypara{MUTANT} \cite{gokhale2020mutant} 
is a state-of-the-art data augmentation method by manipulating images and questions. For example, it applies a GAN-based inpainting network to change the object's color to create extra color questions; it manipulates object numbers using MSCOCO annotations; it masks or negates words to mutate questions.

\mypara{Comparison.} \ourmethod is different from CSS and MUTANT in two aspects. First, CSS and MUTANT can only manipulate already annotated questions for an image, while we can create new questions for an image by borrowing them from other images. Second, CSS needs a pre-trained attention-based VQA model to identify critical objects/words while MUTANT's best version requires additional loss terms for training. In contrast,  \ourmethod is completely detached from model training. 
\subsection{Implementation details}
\label{ss_imple}

\mypara{Data augmentation by \ourmethodbf.}
We speed up the implementation by grouping IQA triplets of the same unique question and only propagating the question once. We remove redundant triplets if the retrieved image already has the same question. To prevent creating too many triplets from paraphrasing (\S~\ref{ss_para}), for each question $q$ we only search for its top-3 similar questions $q'$ and only create $(i', q, a')$ if $a'$ is a rare answer to $q$ --- we define $a'$ to be a rare answer if there are fewer than five $(q, a')$ pairs in the dataset.
\emph{We emphasize that we only apply \ourmethod to IQA triplets in the training set and search images in the training set.}

\mypara{VQA models.} For the base VQA models, we use the released code from corresponding papers. {Please see Appendix for more details.}

\begin{table}[t]
\centering

\footnotesize

\tabcolsep 5pt
\renewcommand\arraystretch{1.0}
\caption{\small \textbf{Statistics on VQA-CP v2 training data.} Miss-answered: the number of \ourmethod examples that a UpDn model trained on the original dataset cannot answer correctly.}
\vspace{-3mm}
\label{tab:stats_simple_aug}
\begin{tabular}{rrrrr}
\toprule
\textbf{\# of samples} & \textbf{All} & \textbf{Y/N} & \textbf{Num} & \textbf{Other} \\
\midrule
Original & 438K & 183K & 52K & 202K \\
\ourmethod & 5,457K & 2,062K & 1,937K & 1,458K \\
Miss-answered & 3,081K & 974K & 1,489K & 618K \\
\bottomrule
\end{tabular}
\vskip-10pt
\end{table}


\begin{table*}[t]
\centering
\small
\tabcolsep 4.5pt
\renewcommand\arraystretch{1.0}
\caption{\small \textbf{Performance on VQA v2 val set and VQA-CP v2 test set.} Our method \ourmethod ({\color{cyan} cyan background}) consistently improves all answer types for different base models on both VQA v2 and VQA-CP. Note that MUTANT (loss)~\cite{gokhale2020mutant} ({\color{gray}gray color}) applies extra loss terms besides data augmentation.}
\label{tab:main}
\vspace{-3mm}
\begin{tabular}{clcccccccc}
\toprule
\multirow{2}{*}{\textbf{Base}} & \multicolumn{1}{c}{\multirow{2}{*}{\textbf{Method}}} & \multicolumn{4}{c}{\textbf{VQA v2 val}} & \multicolumn{4}{c}{\textbf{VQA-CP test}} \\
\cmidrule(r){3-6} \cmidrule(r){7-10}
 & \multicolumn{1}{c}{} & \textbf{All} & \textbf{Y/N} & \textbf{Num} & \textbf{Other} & \textbf{All} & \textbf{Y/N} & \textbf{Num} & \textbf{Other} \\
 \midrule
\multirow{15}{*}{{\rotatebox[origin=c]{0}{UpDn}}} & Baseline~\cite{anderson2018bottom} & 63.48 & 81.18 & 42.14 & 55.66 & 39.74 & 42.27 & 11.93 & 46.05 \\
 & AdvReg~\cite{ramakrishnan2018overcoming} & 62.75 & 79.84 & 42.35 & 55.16 & 41.17 & 65.49 & 15.48 & 35.48 \\

 & RUBi~\cite{cadene2019rubi} & 61.16 & -- & -- & -- & 44.23 & 67.05 & 17.48 & 39.61 \\

 & CF-VQA (SUM)~\cite{niu2020counterfactual} & 63.54 & 82.51 & 43.96 & 54.30 & 53.55 & 91.15 & 13.03 & 44.97 \\
 & SimpleReg~\cite{shrestha2020negative} & 62.60 & -- & -- & -- & 48.90 & 69.80 & 11.30 & 47.80 \\

 & HINT~\cite{selvaraju2019taking} & 63.38 & 81.18 & 42.99 & 55.56 & 46.73 & 67.27 & 10.61 & 45.88 \\
 & SCR+VQA-X~\cite{wu2019self} & 62.20 & 78.80 & 41.60 & 54.50 & 49.45 & 72.36 & 10.93 & 48.02 \\

 & RandImg~\cite{teney2020value} & 57.24 & 76.53 & 33.87 & 48.57 & 55.37 & 83.89 & 41.60 & 44.20 \\
 & Template-based~\cite{kafle2017data} & {63.83} & {81.61} & {41.98} & {56.10} & {39.75} & {43.03} & {14.98} & {44.83} \\
 & CSS~\cite{chen2020counterfactual} & 63.47 & 80.81 & 43.33 & 55.62 & 41.16 & 43.96 & 12.78 & 47.48 \\
 & MUTANT (plain)~\cite{gokhale2020mutant} & -- & -- & -- & -- & 50.16 & 61.45 & 35.87 & 50.14 \\
 & {\color{Gray}{MUTANT (loss)}}~\cite{gokhale2020mutant} & {\color{Gray}{62.56}} & {\color{Gray}{82.07}} & {\color{Gray}{42.52}} & {\color{Gray}{53.28}} & {\color{Gray}{61.72}} & {\color{Gray}{88.90}} & {\color{Gray}{49.68}} & {\color{Gray}{50.78}} \\

 & \cellcolor{LightCyan}\ourmethod (paraphrasing) & \cellcolor{LightCyan}{63.66} & \cellcolor{LightCyan}{81.44} & \cellcolor{LightCyan}{42.56} & \cellcolor{LightCyan}{55.72} & \cellcolor{LightCyan}{52.57} & \cellcolor{LightCyan}{86.56} & \cellcolor{LightCyan}{13.52} & \cellcolor{LightCyan}{45.47} \\

 & \cellcolor{LightCyan}\ourmethod (propagation) & \cellcolor{LightCyan}{64.37} & \cellcolor{LightCyan}{81.91} & \cellcolor{LightCyan}{44.13} & \cellcolor{LightCyan}{56.40} & \cellcolor{LightCyan}{52.27} & \cellcolor{LightCyan}{65.15} & \cellcolor{LightCyan}{45.32} & \cellcolor{LightCyan}{47.42} \\

 & \cellcolor{LightCyan}\ourmethod (propagation + paraphrasing) & \cellcolor{LightCyan}{64.34} & \cellcolor{LightCyan}{81.97} & \cellcolor{LightCyan}{43.91} & \cellcolor{LightCyan}{56.35} & \cellcolor{LightCyan}{52.65} & \cellcolor{LightCyan}{66.40} & \cellcolor{LightCyan}{43.43} & \cellcolor{LightCyan}{47.98} \\
 \midrule

\multirow{8}{*}{{\rotatebox[origin=c]{0}{LMH}}} & Baseline~\cite{clark2019don} & 56.34 & 65.05 & 37.63 & 54.68 & 52.01 & 72.58 & 31.11 & 46.96 \\

 & RMFE~\cite{gat2020removing} & {--} & {--} & {--} & {--} & {54.44} & {74.03} & {49.16} & {45.82} \\

 & CSS~\cite{chen2020counterfactual} & 59.91 & 73.25 & 39.77 & 55.11 & 58.95 & 84.37 & 49.42 & 48.21 \\
 & CSS+CL~\cite{liang2020learning} & 57.29 & 67.27 & 38.40 & 54.71 & 59.18 & 86.99 & 49.89 & 47.16 \\

 & {\color{Gray}{MUTANT (loss)~\cite{gokhale2020mutant}}} & {\color{Gray}{-- }}& {\color{Gray}{--}} & {\color{Gray}{--}} & {\color{Gray}{-- }}& {\color{Gray}{55.38}} & {\color{Gray}{90.99}} & {\color{Gray}{39.74}} & {\color{Gray}{40.99}} \\

 & \cellcolor{LightCyan}\ourmethod (paraphrasing) & \cellcolor{LightCyan}{61.67} & \cellcolor{LightCyan}{78.70} & \cellcolor{LightCyan}{40.21} & \cellcolor{LightCyan}{54.41} & \cellcolor{LightCyan}{53.29} & \cellcolor{LightCyan}{74.12} & \cellcolor{LightCyan}{33.06} & \cellcolor{LightCyan}{47.93} \\

 & \cellcolor{LightCyan}\ourmethod (propagation) & \cellcolor{LightCyan}{62.67} & \cellcolor{LightCyan}{79.24} & \cellcolor{LightCyan}{41.44} & \cellcolor{LightCyan}{55.70} & \cellcolor{LightCyan}{53.58} & \cellcolor{LightCyan}{73.58} & \cellcolor{LightCyan}{37.07} & \cellcolor{LightCyan}{47.63} \\

 & \cellcolor{LightCyan}\ourmethod (propagation + paraphrasing) & \cellcolor{LightCyan}{62.63} & \cellcolor{LightCyan}{79.31} & \cellcolor{LightCyan}{41.71} & \cellcolor{LightCyan}{55.48} & \cellcolor{LightCyan}{53.70} & \cellcolor{LightCyan}{74.79} & \cellcolor{LightCyan}{34.32} & \cellcolor{LightCyan}{47.97} \\
 \midrule

\multirow{7}{*}{{\rotatebox[origin=c]{0}{LXMERT}}} & Baseline~\cite{tan2019lxmert} & {73.06} & {88.30} & {56.81} & {65.78} & {48.66} & {47.49} & {22.24} & {56.52} \\

 & Template-based~\cite{kafle2017data} & {72.30} & {85.36} & {54.47} & {67.10} & {49.63} & {49.96} & {36.33} & {53.10} \\

 & MUTANT (plain)~\cite{gokhale2020mutant} & -- &--  &--  & -- & 59.69 & 73.19 & 32.85 & 59.29 \\
 & {\color{Gray}{MUTANT (loss)~\cite{gokhale2020mutant}}} & {\color{Gray}{70.24}} & {\color{Gray}{89.01}} & {\color{Gray}{54.21}} & {\color{Gray}{59.96}} & {\color{Gray}{69.52}} & {\color{Gray}{93.15}} & {\color{Gray}{67.17}} & {\color{Gray}{57.78}} \\
 
 & \cellcolor{LightCyan}\ourmethod (paraphrasing) & \cellcolor{LightCyan}{74.37} & \cellcolor{LightCyan}{88.78} & \cellcolor{LightCyan}{57.95} & \cellcolor{LightCyan}{67.76} & \cellcolor{LightCyan}{59.09} & \cellcolor{LightCyan}{73.17} & \cellcolor{LightCyan}{28.72} & \cellcolor{LightCyan}{60.04} \\

 & \cellcolor{LightCyan}\ourmethod (propagation) & \cellcolor{LightCyan}{74.96} & \cellcolor{LightCyan}{89.00} & \cellcolor{LightCyan}{60.00} &\cellcolor{LightCyan}{68.25} & \cellcolor{LightCyan}{61.82} & \cellcolor{LightCyan}{68.39} & \cellcolor{LightCyan}{53.35} & \cellcolor{LightCyan}{60.69} \\

 & \cellcolor{LightCyan}\ourmethod (propagation + paraphrasing) & \cellcolor{LightCyan}{74.98} & \cellcolor{LightCyan}{89.04} & \cellcolor{LightCyan}{59.98} & \cellcolor{LightCyan}{68.25} & \cellcolor{LightCyan}{62.24} & \cellcolor{LightCyan}{69.72} & \cellcolor{LightCyan}{53.63} & \cellcolor{LightCyan}{60.69} \\
 \bottomrule
\end{tabular}

\end{table*}

\mypara{Training with \ourmethodbf triplets.} We explore three ways to train with the original ($\mathcal{O}$) triplets and augmented triplets ($\mathcal{A}$). 
The first is to train with both from the beginning ($\mathcal{A}+\mathcal{O}$); 
the second is to train with $\mathcal{O}$ first and then with both ($\mathcal{O}\rightarrow\mathcal{A}+\mathcal{O}$); 
the third is to train with $\mathcal{O}$ first, then with $\mathcal{A}$, and then with $\mathcal{O}$ again ($\mathcal{O}\rightarrow\mathcal{A}\rightarrow\mathcal{O}$).
The rationale of training with multiple stages is to prevent the augmented data from dominating the training process (see \autoref{tab:stats_simple_aug} for the statistics). We note that, there is a huge number of \ourmethod examples that a VQA model trained with $\mathcal{O}$ only cannot answer. Thus, when training with multiple stages, we remove \ourmethod examples that the model can already answer.
We mainly report results using $\mathcal{O}\rightarrow\mathcal{A}\rightarrow\mathcal{O}$, but compare the three ways in \S~\ref{ss_ablation}.
\subsection{Main results on VQA v2 and VQA-CP v2}
\label{ss_main_results}

\autoref{tab:main} summarizes the main results on VQA v2 val and VQA-CP v2 test. 
We experiment \ourmethod with different base VQA models and compare it to state-of-the-art methods. 
\ourmethod achieves consistent gains against the base models on all answer types (columns). When paired with LXMERT, \ourmethod obtains the highest accuracy on both datasets, except MUTANT (loss) which applies extra losses besides data augmentation.  
\mypara{\ourmethodbf improves all answer types.} 
On \textbf{VQA-CP v2}, \ourmethod boosts the overall accuracy of UpDn from $39.74$\% to $52.65$\%, outperforming all but three methods. One key strength of \ourmethod is that it improves all the answer types, including a $\sim$2\% gain on ``Other'' where many methods suffer. Specifically, compared to CF-VQA~\cite{niu2020counterfactual} and RandImg~\cite{teney2020value} which have higher overall accuracy than \ourmethod, \ourmethod outperforms them on the challenging ``Num'' and ``Other''. 
On \textbf{VQA v2}, \ourmethod achieves the highest accuracy using UpDn, improving +0.86\% on ``All'', +0.79\% on ``Yes/No'', +1.77\% on ``Num'', and +0.69\% on ``Other''.
Other methods specifically designed for VQA-CP v2 usually degrade on VQA v2. 

\mypara{\ourmethodbf is model-agnostic.}
\ourmethod can directly be applied to other VQA models. Besides UpDn, in \autoref{tab:main} we show that \ourmethod can lead to consistent gains for two additional VQA models.
LMH is a de-biasing method for UpDn, which however hurts the accuracy on VQA v2. With \ourmethod, LMH can largely improve on VQA v2. LXMERT is a strong transformer-based VQA model, and \ourmethod can also improve upon it,
achieving the highest accuracy on VQA v2 (all answer types) and on VQA-CP v2 (``Other'').

\mypara{Comparison to data augmentation baselines.} \ourmethod notably outperforms the \textbf{template-based method} \cite{kafle2017data}, the closest method to ours. We attribute this to the question diversity via question propagation and paraphrasing. Compared to \textbf{CSS}~\cite{chen2020counterfactual,liang2020learning}, \ourmethod performs better on all answer types on both datasets, using UpDn.
While CSS outperforms \ourmethod on VQA-CP v2 using the de-biasing LMH, its improvement on VQA v2 is smaller than \ourmethod.
Since LXMERT is a general VQA method like UpDn, we expect that \ourmethod will outperform CSS.
Finally, compared to \textbf{MUTANT} \cite{gokhale2020mutant}, \ourmethod achieves better results on VQA-CP v2 against the version without extra loss terms (\ie, MUTANT(plain)).
It is worth noting that while CSS and MUTANT both generate extra data, they cannot improve but degrade on VQA v2 (when using UpDn or LXMERT). In contrast, \ourmethod improves on all cases, suggesting it as a more general data augmentation method for VQA.

\begin{table}[t]
\centering
\small
\tabcolsep 3pt
\renewcommand\arraystretch{1.0}
\caption{\small \textbf{\ourmethodbf (propagation) w/ or w/o verification (cf. \S~\ref{ss_simple_aug_2}) on VQA-CP v2, using the UpDn model.}}
\label{tab:ablation}
\vspace{-3mm}
\begin{tabular}{cccccc}
\toprule
\textbf{Method} & \textbf{Verification} & \textbf{All} & \textbf{Y/N} & \textbf{Num} & \textbf{Other} \\
\midrule
UpDn & -- & 39.74 & 42.27 & 11.93 & 46.05 \\
\midrule
\multirow{2}{*}{\ourmethod} & {\color{gray}{\xmark}} & 51.96 & 64.02 & 44.44 & 47.70 \\
 & \cmark & 52.27 & 65.15 & 45.32 & 47.42 \\
 \toprule
\end{tabular}
\vspace{-3mm}
\end{table}

\subsection{Ablation studies of \ourmethodbf}
\label{ss_ablation}

\mypara{Question propagation vs. paraphrasing.} \ourmethod leverages the original IQA triplets by propagating questions to other images (\S~\ref{ss_simple_aug_2}) or by paraphrasing question using similar questions (\S~\ref{ss_para}). Propagation can ask more questions about an image. For example, the propagated questions in \autoref{fig:1} and \autoref{fig:vqa_qualitative} ask about image contents different from the original questions. In contrast, paraphrasing only paraphrases the original questions of that image.
{As shown in \autoref{tab:main}, question propagation generally leads to better performance, especially on ``Num'' and ``Other'' answers, suggesting the importance of creating additional questions to cover image contents more exhaustively.}

\mypara{On verification for question propagation.}
\autoref{tab:ablation} compares  \ourmethod (propagation) with and without the verification strategies (cf. \S~\ref{ss_simple_aug_2}). 
Verification improves accuracy at nearly all cases.
\mypara{Multiple-stage training.} 
In \autoref{tab:training}, we compare the {three} training strategies with original triplets ($\mathcal{O}$) and augmented triplets ($\mathcal{A}$). We also train on $\mathcal{O}$ for multiple stages (\ie, more epochs) for a fair comparison. $\mathcal{O}\rightarrow\mathcal{A}\rightarrow\mathcal{O}$ in general outperforms others, and we attribute this to the clear separation of clean and noisy data --- the last training stage may correct noisy information learned in early stages~\cite{zhang2021mosaicos}.

{
\mypara{Training with \ourmethodbf triplets alone.}
We further investigate training the UpDn model with augmented triplets alone ($\mathcal{A}$). On VQA-v2, we get 39.62\% overall accuracy, worse than the baseline trained with original data (63.48\%). This is likely due to the noise in the augmented data. On VQA-CP, we get 51.60\%, much better than the baseline (39.74\%) but worse than training with both augmented and original triplets (52.65\%). We surmise that \ourmethod triplets help mitigate the language bias shifts in VQA-CP. Please see Appendix for a human study on assessing the quality of the triplets augmented by \ourmethod.
}

\begin{table}[t]
\centering
\small
\tabcolsep 2.8pt
\renewcommand\arraystretch{1.0}
\caption{\small \textbf{A comparison of training strategies on VQA-CP v2 with the UpDn model.}  $\mathcal{O}$: original triplets. $\mathcal{A}$: augmented triplets by \ourmethod.
}
\label{tab:training}
\vspace{-3mm}
\begin{tabular}{cccccc}
\toprule
\textbf{Method} & \textbf{Strategy} & \textbf{All} & \textbf{Y/N} & \textbf{Num} & \textbf{Other} \\
\midrule
\multirow{2}{*}{UpDn} & $\mathcal{O}$ & 39.74 & 42.27 & 11.93 & 46.05 \\
 & $\mathcal{O}\rightarrow\mathcal{O}\rightarrow\mathcal{O}$ & 39.47 & 43.11 & 11.75 & 45.16 \\
 \midrule
\multirow{3}{*}{\ourmethod} & $\mathcal{A}$ + $\mathcal{O}$ & 47.50 & 59.76 & 38.18 & 43.63 \\
 & $\mathcal{O}\rightarrow\mathcal{A}$ + $\mathcal{O}$ & 49.73 & 59.67 & 36.58 & 48.12 \\
 & $\mathcal{O}\rightarrow\mathcal{A}\rightarrow\mathcal{O}$ & 52.65 & 66.40 & 43.43 & 47.98 \\
 \bottomrule
\end{tabular}
\end{table}

{
\mypara{Effects of augmentation types.}
We experiment with propagating each question type alone on VQA-CP v2, using UpDn as the base model. In \autoref{tab:separate}, we show the separate results of \ourmethod with different question types. The augmented questions notably improve the corresponding answer type.}

\begin{table}[t]
\centering
\small
\tabcolsep 4pt
\renewcommand\arraystretch{1.0}
\caption{\small {\textbf{Effects of different augmention types (cf. \S~\ref{ss_simple_aug_2}).} We report results on VQA-CP v2, using the UpDn model.}}
\label{tab:separate}
\vspace{-3mm}
\begin{tabular}{cccccc}
\toprule
\textbf{Method} & \textbf{Aug Type} & \textbf{All} & \textbf{Y/N} & \textbf{Num} & \textbf{Other} \\
\midrule
UpDn & -- & 39.74 & 42.27 & 11.93 & 46.05 \\
\midrule
\multirow{5}{*}{\ourmethod} & Y/N & 47.20 & \textbf{68.63} & 12.12 & 45.68 \\
 & Num & 44.62 & 42.87 & \textbf{43.80} & 45.77 \\
 & Color & 40.97 & 43.11 & 12.39 & 47.68 \\
 & Other & 41.22 & 43.17 & 12.19 & \textbf{48.16} \\
 & All & \textbf{52.65} & 66.40 & 43.43 & 47.98 \\
 \bottomrule
\end{tabular}
\end{table}

\subsection{\ourmethodbf in additional scenarios}
\label{ss_semi_sup}

We explore \ourmethod in the scenarios where there are (i) limited questions per image, and (ii) extra weakly-labeled or unlabeled images. For (ii), both have no IQA triplets but the weakly-labeled ones have human-annotated object instances.

\mypara{Learning with limited triplets.}
{We randomly keep a fraction of annotated QA pairs for each training image on VQA-CP v2. \autoref{tab:weakly_question} shows that even under this annotation-scarce setting (\eg, only 10\% of QA pairs are kept), \ourmethod can already be effective, outperforming the baseline UpDn model trained with all data. This demonstrates the robustness of \ourmethod on dealing with the challenging setting with limited triplets.}

\begin{table}[t]
\centering
\small
\tabcolsep 4pt
\renewcommand\arraystretch{1.0}
\caption{\small {\textbf{Learning with limited IQA triplets on VQA-CP v2.} We keep a certain fraction of QA pairs per image.}}
\label{tab:weakly_question}
\vspace{-3mm}
\begin{tabular}{cccccc}
\toprule
\multirow{1}{*}{\textbf{Method}} & \multirow{1}{*}{\textbf{Fraction}}  &\textbf{All} & \textbf{Y/N} & \textbf{Num} & \textbf{Other} \\
 \midrule
\multirow{1}{*}{UpDn} & 1.00  & 39.74 & 42.27 & 11.93 & 46.05 \\
\midrule
\multirow{4}{*}{\ourmethod} & 1.00  & {52.65} & {66.40} & {43.43} & {47.98} \\
 & 0.50  & 47.67 & 57.65 & 37.77 & 45.17 \\
 & 0.25 & 46.03 & 52.01 & 37.96 & 45.12 \\
 & 0.10 & 42.91 & 45.09 & 30.24 & 45.25 \\
\bottomrule
\end{tabular}
\vspace{-3mm}
\end{table}

\mypara{Learning with weakly-labeled or unlabeled images.}
We simulate the scenarios by keeping the QA pairs for a fraction of images (\ie, labeled data) and removing the QA pairs entirely for the other images.
Conventionally, a VQA model cannot benefit from the images without QA pairs, but \ourmethod could leverage them by propagating questions to them. Specifically, for
images without QA pairs, we consider two cases. We either keep their MSCOCO object instance annotations (\ie, weakly-labeled data) or completely rely on object detectors (\ie, unlabeled data). 
\autoref{tab:semi_image} shows the results, in which we only apply \ourmethod to the weakly-labeled and unlabeled images. As shown, \ourmethod yields consistent improvements, opening up the possibility of leveraging additional images to improve VQA.

\begin{table}[t]
\centering
\small
\tabcolsep 3.5pt
\renewcommand\arraystretch{1.0}
\caption{\small \textbf{Learning with weakly-labeled and unlabeled images for VQA v2.} Fraction: the portion of images with annotated QA pairs. GT: MSCOCO ground truth annotations. OD: Faster R-CNN object detection. {\color{gray}{\xmark}}: supervised training with only labeled VQA training examples.}
\vspace{-3mm}
\label{tab:semi_image}
\begin{tabular}{cccccc}
\toprule
\multirow{1}{*}{\textbf{Fraction}} & \multirow{1}{*}{\textbf{\ourmethodbf}} & \textbf{All} & \textbf{Y/N} & \textbf{Num}& \textbf{Other} \\
 \midrule
\multirow{2}{*}{1.00} & {\color{gray}{\xmark}} & 63.48 & 81.18 & 42.14 & 55.66 \\
 & \cmark & 64.34 & 81.97 & 43.91 & 56.35 \\
 \midrule
\multirow{3}{*}{0.50} & {\color{gray}{\xmark}} & 60.93 & 78.45 & 40.74 & 52.96 \\
 & GT & 61.47 & 78.92 & 41.43 & 53.50 \\
 & OD & 61.47 & 78.93 & 41.42 & 53.50 \\
 \midrule
\multirow{3}{*}{0.25} & {\color{gray}{\xmark}} & 56.70 & 74.02 & 37.81 & 48.53 \\
 & GT & 57.54 & 74.49 & 39.08 & 49.54 \\
 & OD & 57.56 & 74.63 & 38.67 & 49.57  \\
 \midrule
\multirow{3}{*}{0.10} & {\color{gray}{\xmark}} & 51.06 & 69.18 & 33.46 & 41.93 \\
 & GT & 52.18 & 69.76 & 35.95 & 43.10 \\
 & OD & 52.27 & 69.98 & 35.07 & 43.34 \\
 \bottomrule
\end{tabular}
\end{table}

\subsection{Qualitative results}
\label{ss_qual}

\begin{figure}[t]
    \centerline{\includegraphics[width=1\linewidth]{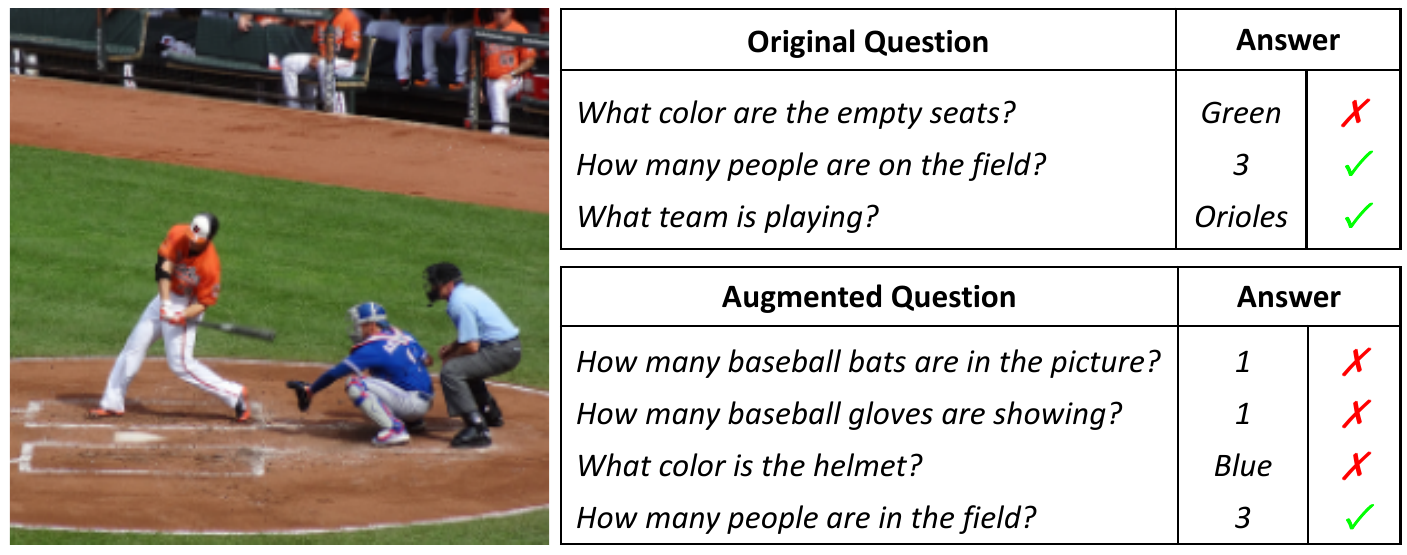}}
    \caption{\small {\textbf{Qualitative results.} We show the training image and its QA pairs from VQA-CP, and the generated QA pairs by \ourmethod. {\color{green}{\cmark}}/{\color{red}{\xmark}} indicates if the baseline VQA model (trained without \ourmethod) answers correctly/incorrectly. In augmented QA pairs, the first three are from question propagation and the last one is by paraphrasing.}}
    \label{fig:vqa_qualitative}
    \vspace{-3mm}
\end{figure}

We show a training image and its augmented QA pairs by \ourmethod in \autoref{fig:vqa_qualitative}. A VQA model trained on the original IQA triplets cannot answer many of the newly generated questions, even if the image is in the training set, showing the necessity to include them for training a stronger model. {More qualitative results can be found in Appendix.}

\section{Conclusion}
\label{s_disc}
We proposed \ourmethod, a data augmentation method for VQA that can turn information already in the datasets into explicit IQA triplets for training. 
\ourmethod is simple but by no means trivial. First, it justifies that mid-level vision tasks like object detection can effectively benefit VQA. Second, we probably will never be comprehensive enough in annotating data, and \ourmethod can effectively turn what we have at hand (\ie, ``knowns'') to examples a VQA model wouldn't have known (\ie, ``unkowns''). \ourmethod can notably improve the accuracy of VQA models on both VQA v2~\cite{goyal2017making} and VQA-CP v2~\cite{clark2019don}.

{
\mypara{Acknowledgements.}
This research is partially supported by NSF IIS-2107077 and the OSU GI Development funds.
We are thankful for the generous support of computational resources by the Ohio Supercomputer Center.
}

\bibliographystyle{acl_natbib}
\bibliography{main}
\clearpage
\appendix

\begin{strip}
\centering
	\textbf{\Large Discovering the Unknown Knowns: \\ Turning Implicit Knowledge in the Dataset into Explicit Training Examples for Visual Question Answering \\[0.2em] {\Large (Appendix)}}
	\vskip 10pt
\author{Jihyung Kil, \hfill Cheng Zhang, \hfill Dong Xuan, \hfill Wei-Lun Chao \\
The Ohio State University, Columbus, OH, USA \\
\small \texttt{\{kil.5, zhang.7804, xuan.3, chao.209\}@osu.edu}
}
\vspace{10pt}
\end{strip}

In this appendix, we provide details and results omitted in the main text.
\begin{itemize}[leftmargin=12pt]
    \item \autoref{app_imple}: additional implementation details. (\S~\ref{ss_imple} of the main paper)
    
    \item \autoref{app_gqa}: results on GQA dataset. (\S~\ref{ss_setup} of the main paper)
    
    \item \autoref{human_study}: human study for assessing the quality of triplets generated by \ourmethod. (\S~\ref{ss_ablation} of the main paper)
    
    \item \autoref{qual_app}: additional qualitative results generated by \ourmethod. (\S~\ref{ss_qual} of the main paper)
\end{itemize}

\section{Additional Implementation Details}
\label{app_imple}

\subsection{Baseline VQA models.}
We validate \ourmethod with three VQA models in our experiments: Bottom-Up Top-Down (UpDn)\footnote{UpDn model implementation: \url{https://github.com/yanxinzju/CSS-VQA}.}~\cite{anderson2018bottom}, Learned-Mixin+H (LMH)\footnote{LMH model implementation: \url{https://github.com/chrisc36/bottom-up-attention-vqa}.}~\cite{clark2019don}, and LXMERT\footnote{LXMERT model implementation: \url{https://github.com/airsplay/lxmert}.}~\cite{tan2019lxmert}. All baseline models are implemented using officially released codebase. More details of code and data are publicly available at \url{https://github.com/heendung/simpleAUG}.

\subsection{Optimization}
\label{app_opt}

\mypara{UpDn and LMH.}
We maintain the default settings in UpDn and LMH except for using the mini-batch size of $512$ on VQA v2 and $1,024$ on VQA-CP v2. Following the official implementations, our visual features are the output of Faster R-CNN~\cite{ren2016faster} object detector trained on Visual Genome~\cite{krishna2017visual}, provided by \citet{anderson2018bottom}. We optimize UpDn and LMH using stochastic gradient descent (SGD) with Adamax~\cite{kingma2014adam} and learning rate $2\times10^{-4}$. Training a baseline UpDn or LMH model on a single NVIDIA RTX A6000 takes around 2 hours for convergence.

\mypara{LXMERT.}
Similar to UpDn and LMH models, LXMERT leverages the object features from the Faster R-CNN detection provided by \citet{anderson2018bottom}. We train a LXMERT model using the mini-batch size of $256$. Following \citet{tan2019lxmert}, we use Adam~\cite{kingma2014adam} as the optimizer with a linear decayed learning rate schedule. Training a baseline LXMERT model on a single NVIDIA RTX A6000 takes around 8 hours for convergence.

\mypara{Multi-stage training.}
As discussed in \S~\ref{ss_imple} and \S~\ref{ss_ablation} of the main paper,
we train the VQA models with a three-stage paradigm ($\mathcal{O}\rightarrow\mathcal{A}\rightarrow\mathcal{O}$): first with original triplets $\mathcal{O}$, then with the \ourmethod triplets $\mathcal{A}$, and then with $\mathcal{O}$ again. \emph{In each of these three stages, we follow the same optimization procedures as we train the baseline VQA models in the first stage.} We report the best results on VQA v2 validation set~\cite{goyal2017making} and VQA-CP v2 test set~\cite{agrawal2018don}.

\subsection{Additional details of \ourmethodbf}
As mentioned in the main paper, for each annotated IQA triplet $(i,q,a)$ in the dataset, \ourmethod propagates $q$ to other relevant images. To begin with, we find unique questions by filtering out any duplicate sentences. We then extract meaningful words from the unique questions in line with~\cite{chen2020counterfactual,wu2019self}. Concretely, we remove the question type from $q$ and then apply a spaCy part-of-speech (POS) tagger~\cite{honnibal2017spacy} to extract ``nouns''. To handle the synonyms, we further consider the singular/plural forms and super-categories of nouns\footnote{Paraphrase database~\cite{ganitkevitch2013ppdb} or WordNet~\cite{miller1995wordnet} could be used to handle other synonyms.}. Moreover, we remove non-informative words (\eg, ``picture'' or ``photo'') in the sentence. For example, in a question, ``What is the man doing in the picture?'', the word ``picture'' refers to an image itself but not any specific object. 
There are around 8\% of triplets like this in VQA v2~\cite{goyal2017making}. While it is possible that both sentence and image may contain such non-informative contents (\eg, ``How many pictures on the wall?''), there are $<$1\% such questions.

\section{Results on GQA Dataset}
\label{app_gqa}

We further conduct a preliminary study of \ourmethod on the popular GQA dataset~\cite{hudson2019gqa}, which focuses on compositional VQA tasks and consists of 22M questions about various day-to-day images. Each image in GQA is associated with a scene graph~\cite{johnson2015image} which consists of the objects, attributes, and relationships.

We focus on binary questions ($35$\% of all questions) and propagate a question $q$ to an image $i$ according to the image’s scene graph. Particularly, we leverage the semantic type of the question (\eg, ``attribute'', ``relation'') and the scene graph to generate the answer. For example, suppose $q$ asks if an object contains a certain ``attribute'', we check the scene graph's node of that object to determine the answer. \ourmethod can improve the accuracy of UpDn from $56.06$\% to $56.52$\%, justifying its generalizability and applicability.

\section{Human Evaluation on \ourmethodbf Triplets}
\label{human_study}

\begin{figure*}[t]
    \centerline{\includegraphics[width=1\linewidth]{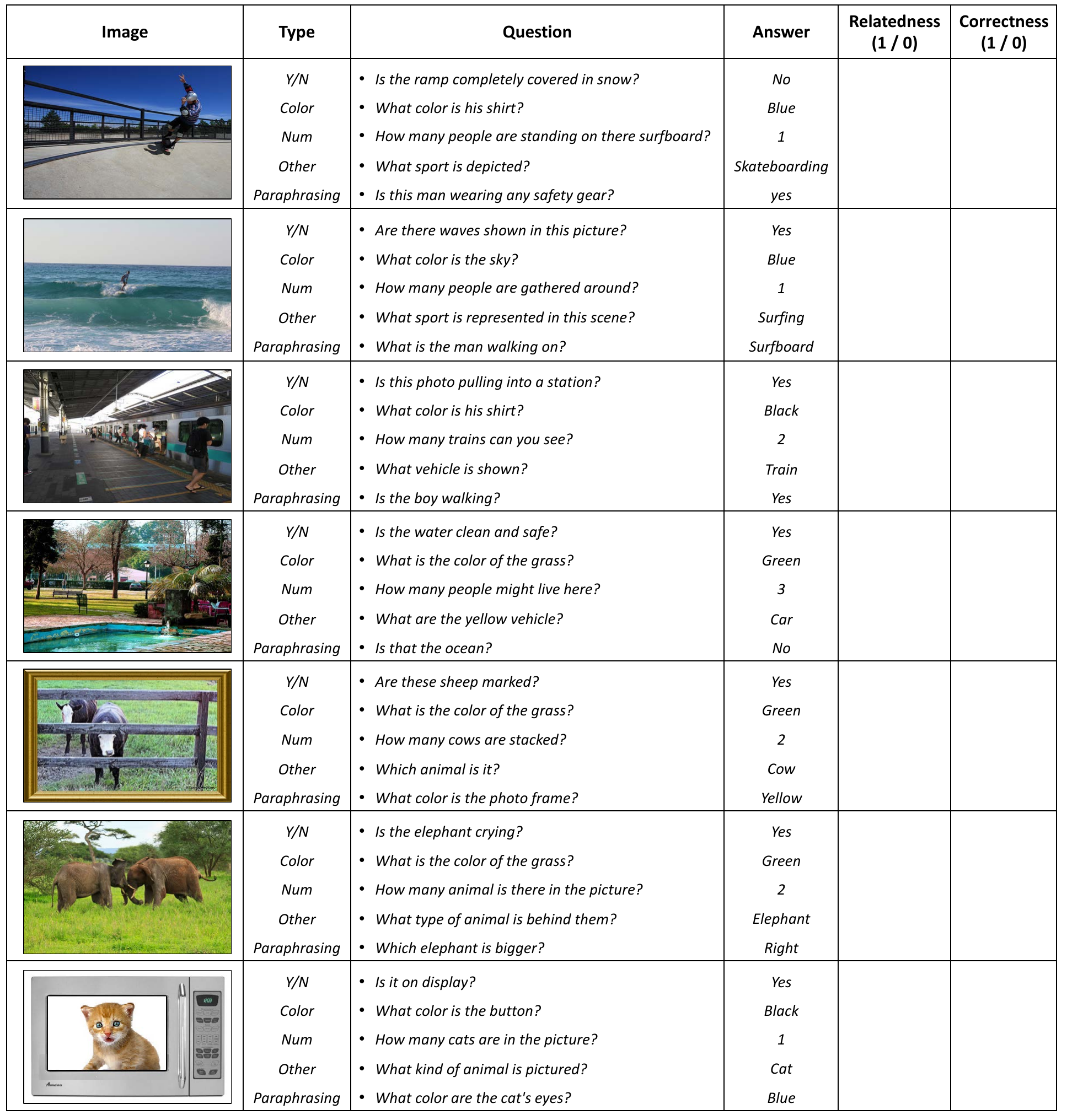}}
    \caption{\small \textbf{Examples in human study.} For each image, we pick 5  triplets created by \ourmethod (4 by propagation Y/N, Num, Other, Color and 1 by paraphrasing) and ask crowd workers to evaluate the IQA triplets by the question’s \emph{relatedness (1 / 0)} to the image and the answer’s \emph{correctness (1 / 0)} to the image and question. }
    \label{fig:human_study}
\end{figure*}

\ourmethod requires no sentence/image generation steps, and thus all examples are natural annotations from humans, largely alleviating the artificial noise that the previous methods
may have. To further evaluate the quality of the augmented triplets, we randomly select $500$ images and pick $5$ augmented QA pairs per image from each type ($4$ by propagation {Y/N, Num, Other, Color} and $1$ by paraphrasing). For those $2,500$ triplets, we ask $5$ different crowd workers to evaluate ``relatedness ($1/0$)'' of the augmented questions and ``correctness ($1/0$)'' of the answer given the question and image. That is, if the question makes sense for the corresponding image, rate $1$, otherwise $0$; if the answer is correct, rate $1$, otherwise $0$ (see \autoref{fig:human_study}). \autoref{tab:human_eval} shows the human study results. The average relatedness / correctness are $86.75$\% / $67.60$\% for propagation and $80.80$\% / $64.40$\% for paraphrasing. 

We note that these generated data are based on human-annotated questions in the dataset. Therefore, there are no artifacts in the questions. Moreover, these generated data are to augment the original data. Thus, even if they contain noise, they can consistently improve the model’s performance. 

\begin{table}[t]
\centering
\small
\tabcolsep 1pt
\renewcommand\arraystretch{1.0}
\caption{\small \textbf{Human evaluation.} The Relatedness and Correctness are shown on different types / question types.}
\label{tab:human_eval}
\vspace{-3mm}
\begin{tabular}{cccc}
\toprule
\multirow{1}{*} {\ourmethodbf} & \textbf{Type} & \textbf{Relatedness (\%)} & \textbf{Correctness (\%)} \\
\midrule
\multirow{5}{*} & Y/N & 82.60 & 52.20 \\
& Color & 87.20 & 77.20 \\
Propagation & Num & 89.20 & 60.80 \\
& Other & 88.00 & 80.20 \\
\cmidrule{2-4}
& Overall & 86.75 & 67.60 \\
\midrule
\multirow{1}{*} {Paraphrasing} & Overall & 80.80 & 64.40 \\
\bottomrule
\end{tabular}
\end{table}

\section{Additional Qualitative Results by \ourmethodbf}
\label{qual_app}

\begin{figure*}[h]
    \centerline{\includegraphics[width=1\linewidth]{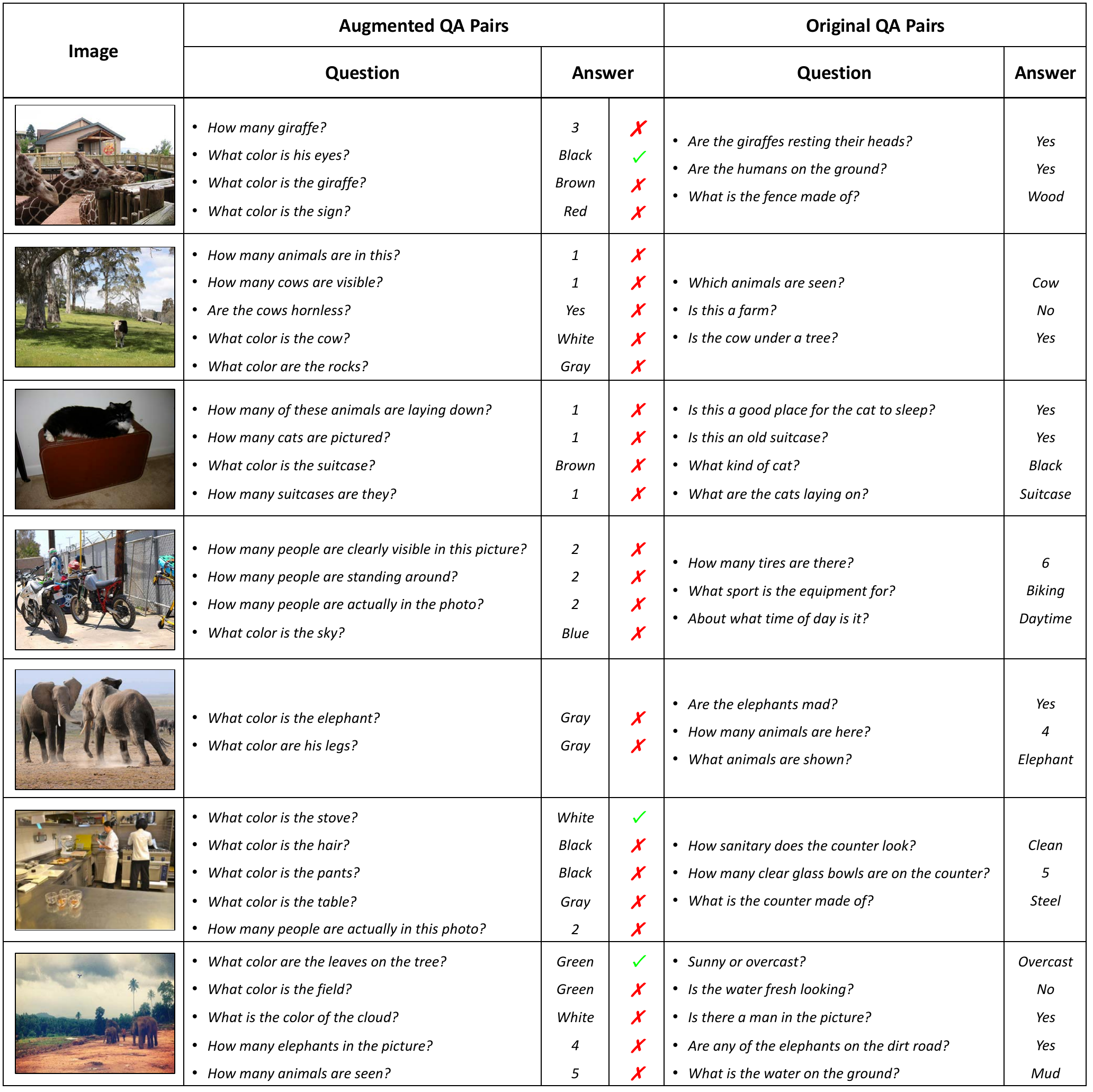}}
    \caption{\small \textbf{Additional qualitative results on VQA-CP.} We show the original image, the generated QA pairs by \ourmethod, and the original QA pairs. {\color{green}{\cmark}}/{\color{red}{\xmark}} indicates if the baseline VQA model (trained without \ourmethod) predicts correctly/incorrectly.}
    \label{fig:qual_app}
\end{figure*}

\autoref{fig:qual_app} provides more qualitative results. We note that the baseline model has still suffered in learning many IQA relationships. Even for simple cases (\eg, What color is the elephant?), the model is not able to answer them correctly. Thus, augmenting IQA triplets with the implicit information in the dataset can notably improve the model's performance on both VQA v2 and VQA-CP v2.

\end{document}